\documentclass[letterpaper, 10 pt, conference]{ieeeconf}  

\IEEEoverridecommandlockouts                              

\usepackage{cite}
\usepackage{amsmath,amssymb,amsfonts}
\usepackage{algorithmic}
\usepackage{graphicx}
\usepackage{textcomp}
\usepackage[table]{xcolor}
\usepackage{adjustbox}
\usepackage{url}
\usepackage{booktabs}
\usepackage{caption}
\usepackage{forest}
\usepackage{dirtree}

\title{\LARGE \bf
DriveIndia: An Object Detection Dataset for Diverse \\Indian Traffic Scenes
}

\author{Rishav Kumar, D. Santhosh Reddy, and P. Rajalakshmi
\thanks{Corresponding author: Rishav Kumar (ai22mtech12003@iith.ac.in)}%
}

\begin{document}

\maketitle
\thispagestyle{empty}
\pagestyle{empty}


\begin{abstract}

We introduce \textbf{DriveIndia}, a large-scale object detection dataset purpose-built to capture the complexity and unpredictability of Indian traffic environments. The dataset contains \textbf{66,986 high-resolution images} annotated in YOLO format across \textbf{24 traffic-relevant object categories}, encompassing diverse conditions such as varied weather (fog, rain), illumination changes, heterogeneous road infrastructure, and dense, mixed traffic patterns and collected over \textbf{120+ hours} and covering \textbf{3,400+ kilometers} across urban, rural, and highway routes. DriveIndia offers a comprehensive benchmark for real-world autonomous driving challenges. We provide baseline results using state-of-the-art \textbf{YOLO family models}, with the top-performing variant achieving $mAP_{50}$ of \textbf{78.7\%}. Designed to support research in robust, generalizable object detection under uncertain road conditions, DriveIndia will be publicly available via the TiHAN- IIT Hyderabad dataset repository (\url{https://tihan.iith.ac.in/TiAND.html}).

\end{abstract}

\section{Introduction}

Autonomous driving systems rely heavily on robust object detection to perceive and navigate complex environments. While significant progress has been made using deep learning, particularly convolutional neural networks (CNNs), most public benchmarks such as KITTI \cite{b5} and nuScenes \cite{nuscenes} are designed around structured, rule-compliant traffic scenarios common in Western countries. However, these conditions differ dramatically from those encountered in regions like India, where unstructured traffic, variable infrastructure, and environmental uncertainty are the norm.

In India, road conditions are uniquely challenging. Urban and rural areas often lack standardized signage or lane markings, while traffic comprises a heterogeneous mix of vehicles, pedestrians, animals, and non-motorized transport. These factors make India an ideal testbed for developing generalizable perception models that can handle uncertainty and real-world complexity.

Several Indian datasets have emerged to address local challenges, such as the Indian Driving Dataset (IDD) \cite{b1, IDD-2023}, DATS \cite{b4}, NITCAD \cite{b2}, and the recent multimodal TIAND dataset \cite{tiand}. While these provide valuable resources, they often focus on specific tasks (e.g., semantic segmentation), limited object classes, or urban-centered use cases, leaving a gap in large-scale, diverse object detection datasets covering varied road types and environmental conditions.

\begin{figure} \centering \includegraphics[width=\linewidth]{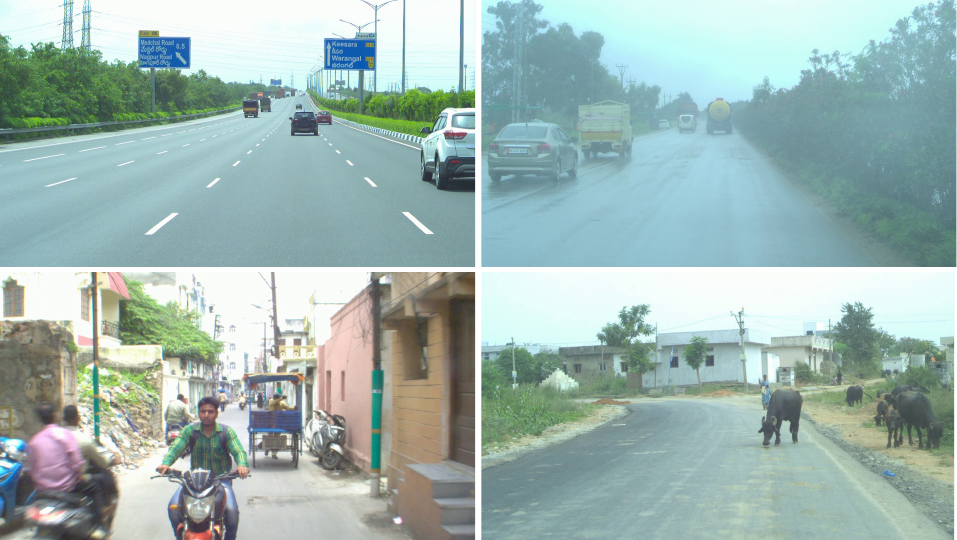} \caption{Sample scenes from the DriveIndia dataset: urban roads, rural highways, fog, and rain.} \label{fig:Main_image} \end{figure}

To address this gap, we introduce \textbf{DriveIndia} (refer Fig. \ref{fig:Main_image}), a large-scale object detection dataset explicitly tailored for Indian road environments. It contains 66,986 high-resolution RGB images annotated in the YOLO format, with normalized bounding boxes stored in per-image .txt files. The annotations span 24 traffic-relevant object classes. The data was captured under diverse conditions—including fog, rain, low light, dense traffic, and infrastructure anomalies—making DriveIndia well-suited for training models that need to perform reliably in unstructured and unpredictable real-world scenarios. All images are recorded at a resolution of 1920×1080 pixels, ensuring clear visual quality for object localization and model training.

While prior work such as TIAND \cite{tiand} focuses on multimodal autonomy using LiDAR and radar sensors, DriveIndia emphasizes general-purpose visual object detection across geographic, environmental, and traffic diversity. It further provides a benchmark for evaluating real-time detectors using YOLO-series models \cite{yolo}, establishing new baselines for Indian road conditions.

\subsection*{Key Contributions} 
\begin{itemize} 
    \item Developed \textbf{DriveIndia}, the largest publicly available object detection dataset focused on unstructured Indian traffic, featuring $\sim$67,000 annotated images. 
    \item It covers 24 traffic-relevant object classes labeled in YOLO format with normalized bounding boxes.
    \item We establish strong detection baselines using state-of-the-art models.
    
    \item The dataset will be publicly released via the TiHAN-IIT Hyderabad portal \cite{tihanportal} to support open research.
\end{itemize}

\begin{table*}[t]
\caption{Comparison of major Indian and global object detection datasets. DriveIndia offers the largest volume of annotated images under uncertain weather conditions (fog, rain), while IDD and DATS focus on structured urban scenes.}
\begin{center}
\resizebox{\textwidth}{!}{%
\renewcommand{\arraystretch}{1.3}
\begin{tabular}{|l|c|c|c|c|c|c|}
\hline
\textbf{Attribute} & \textbf{DriveIndia (Ours)} & \textbf{IDD} & \textbf{DATS}  & \textbf{ITD} & \textbf{BDD100K} \\
\hline
\textbf{Release Year} & 2025 & 2023 & 2022 & 2020 & 2020 \\
\textbf{Task Type} & Object Detection & 3D Object Detection & Object Detection & Vehicle Classification & Detection / Segm. / Tracking \\
\textbf{Images} & 66,986 & 46,588 & 10,000+ & 17,666 (9.8k from IDD) & 100,000 \\
\textbf{Object Classes} & 24 & 17 \textit{(10 used)} & 45 & ~10 & 10 \\
\textbf{Camera Type} & Vehicle-mounted & Vehicle-mounted & Vehicle-mounted & Fixed (CCTV) & Vehicle-mounted \\
\textbf{Environment} & Urban, Rural, Highway & Urban, Suburban & Urban, Rural & Junction-level & All (urban, rural) \\
\textbf{Weather Conditions} & Fog, Rain & Clear only & Clear only & Clear only & Fog, Rain, Night \\
\textbf{Night-time Data} & No & No & Yes & No & Yes \\
\textbf{Annotation Format} & TXT (YOLO-style) & COCO (JSON) & XML / TXT / JSON & Class Labels & COCO / MOT / Video \\
\hline
$\mathbf{mAP_{50}}$& 78.7\% & 67.5\% & Not Reported & 53.9\% & 56.7\% \\
$\mathbf{mAP_{50{:}95}}$ & 51.7\% & 33\% & Not Reported & 42\% & 39\% \\
\hline
\end{tabular}%
}
\label{tab:dataset_comparison}
\end{center}
\end{table*}

We hope DriveIndia is a foundational resource for advancing perception models in diverse, real-world traffic environments, particularly for emerging markets.


\section{Related Work and Dataset Comparison}
Object detection in real-world driving scenarios has advanced rapidly with the release of large-scale datasets. However, most public benchmarks such as KITTI, Cityscapes \cite{cityscapes}, and BDD100K \cite{bdd100k} are designed around structured, rule-compliant traffic environments typically found in developed countries. These datasets, while valuable, fall short of representing the unstructured, heterogeneous, and often chaotic nature of traffic conditions prevalent in countries like India.

In recent years, several Indian datasets have emerged to address this gap. The Indian Driving Dataset (IDD) \cite{b1}, initially released in 2018, focuses on semantic segmentation in unstructured roads, particularly in urban and suburban environments. A more recent extension, IDD \cite{IDD-2023}, includes bounding box annotations for object detection across 46,588 images, but does not cover adverse weather conditions.

DATS \cite{b4} was curated to support object detection in dense urban Indian traffic and includes critical classes such as autorickshaws and pedestrians. Although it annotates in XML, TXT, and COCO formats, it is geographically constrained and lacks diversity in environmental settings.

The Indian Traffic Dataset (ITD) \cite{b3} from IIT Roorkee combines 9,888 images from IDD with 7,778 newly collected ones, focusing on vehicle classification for adaptive traffic control. It uses fixed surveillance-style camera setups and does not provide detailed bounding box annotations suitable for mobile perception systems.

For global comparison, BDD100K \cite{bdd100k} remains the most comprehensive benchmark with 1,00,000 images covering object detection, semantic segmentation, and tracking across varied weather, time-of-day, and road conditions. While it includes night-time and rainy scenarios, it lacks the localized diversity of road users, signage, and informal traffic behavior in Indian settings.

In addition to dataset-focused efforts, several studies have explored the performance of deep learning models on Indian road scenes. Mukhopadhyay et al. \cite{b6} presented a comparative analysis of different CNN architectures for object detection on localized road datasets, highlighting model selection and feature extraction challenges. Similarly, Sehgal et al. \cite{b7} conducted a comparative study of deep learning models for vehicle detection in unconstrained Indian road environments, underscoring the performance limitations of standard architectures in chaotic, heterogeneous traffic conditions. While these works provide valuable insights into model behavior, they primarily focus on architectural comparisons rather than addressing the broader need for large-scale, diverse datasets covering multiple object classes and environmental conditions, which DriveIndia aims to fulfill.

Compared to these, DriveIndia offers a significantly larger and more diverse set of object annotations, covering challenging scenarios such as fog, rain, and unstructured rural roads. It enables benchmarking of real-time object detectors under conditions typical of developing regions. A detailed comparison is provided in Table~\ref{tab:dataset_comparison}.

\section{DriveIndia Dataset Overview}

DriveIndia is a large-scale object detection dataset created to reflect the complexity and uncertainty of real-world traffic conditions in India. It consists of \textbf{66,986} high-resolution RGB images, split into 53,586 for training, 6,700 for validation, and 6,700 for testing, captured over 120+ hours of driving and covering more than 3,400 kilometers across highways, urban centers, and rural regions. The dataset comprises a total of approximately \textbf{4,71,092} annotated instances across 24 traffic-relevant object categories. Data was recorded using vehicle-mounted cameras operating at 1920×1080 resolution, providing detailed visual context for object detection tasks.

This section outlines the data collection methodology, annotation process, and environmental diversity represented in the dataset.

\subsection{Data Collection Strategy}

\begin{figure}[h]
    \centering
    \includegraphics[width=0.71
    \linewidth]{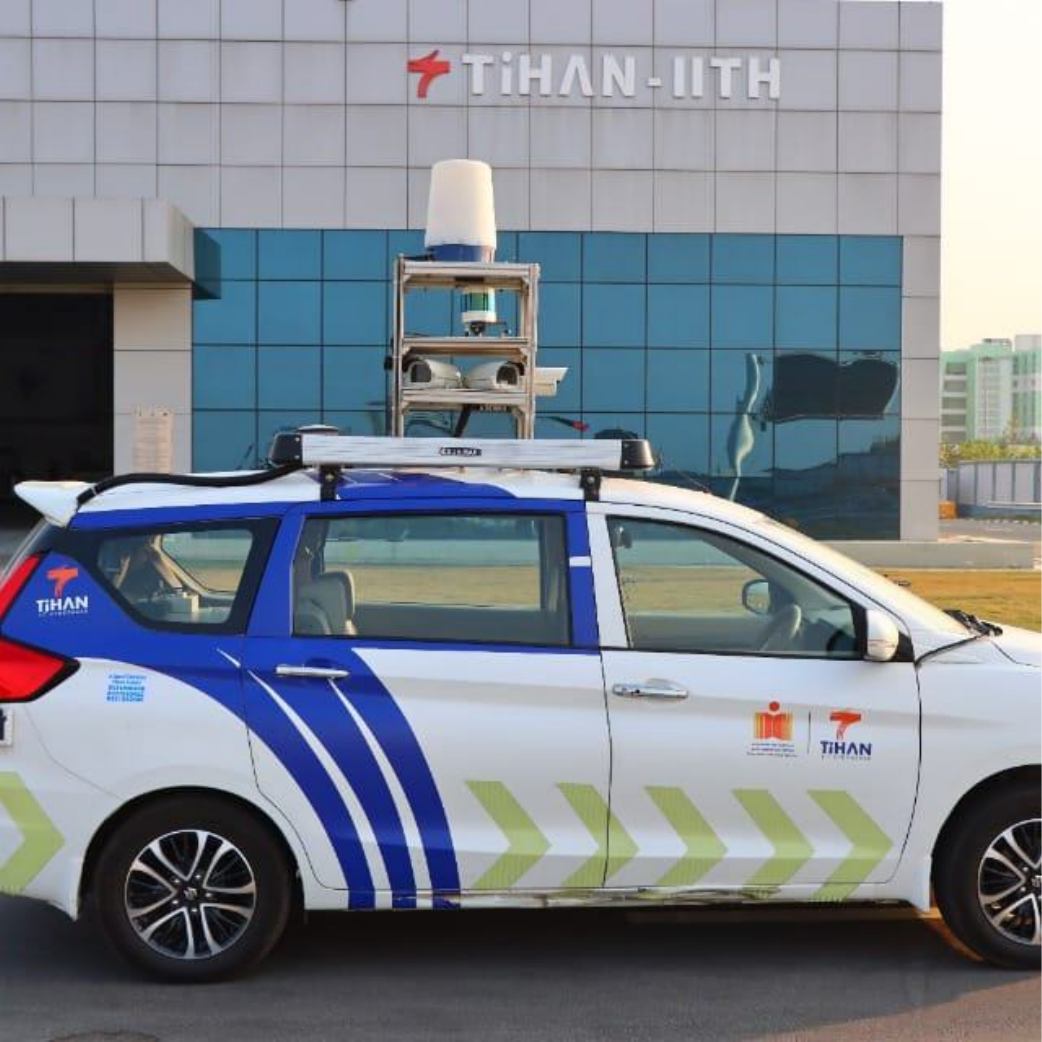}
    \caption{Data Collection Vehicle with Front \& Back Camera}
    \label{fig:data-collection-vehicle}
\end{figure}

We collected data across a wide geographic spread in southern India, spanning highways, city streets, rural roads, and construction areas beyond Hyderabad. The recording setup used multiple synchronized RGB cameras mounted on passenger vehicles (see Fig.~\ref{fig:data-collection-vehicle}).

The collection periods were carefully chosen to span different times of day and to include a wide range of environmental scenarios. 

\subsection{Annotation Format and Tools}

All images in DriveIndia were annotated using the YOLO format, where each object instance is represented as a single line in a per-image `.txt' file.

Annotation was performed using a combination of LabelImg \cite{labelImg} and custom validation scripts to ensure consistency and accuracy.

\begin{figure}[t]
    \centering
    \includegraphics[width=\linewidth]{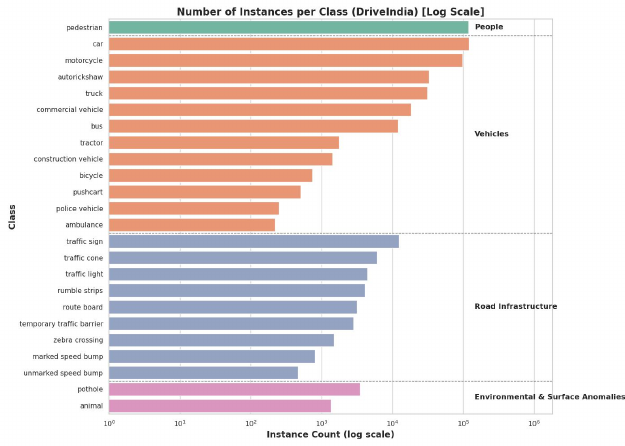}
    \caption{Distribution of class labels in the DriveIndia Dataset}
    \label{fig:classes}
\end{figure}

\textbf{The 24 object classes included in DriveIndia are grouped} as follows see Fig.~\ref{fig:classes} for their frequency distribution:

\begin{itemize}
    \item \textbf{Person} : Pedestrian
    \item \textbf{Vehicles}: Bicycle, Car, Motorcycle, Bus, Commercial vehicle, Truck, Autorickshaw, Ambulance, Police vehicle, Tractor, Pushcart, Construction vehicle
    \item \textbf{Road Infrastructure}: Route board, Traffic sign, Traffic light, Temporary traffic barrier, Traffic cone, Rumblestrips, Unmarked speed bump, Marked speed bump, Zebra crossing
    \item \textbf{Environmental \& Surface Anomalies}: Animal, Pothole
\end{itemize}

These categories were selected to reflect the vast diversity of road users and objects commonly encountered in Indian traffic environments, including region-specific classes not present in most international datasets. 

\subsection{Weather and Scene Diversity}

One of DriveIndia’s distinguishing features is its deliberate inclusion of challenging environmental conditions rarely addressed in mainstream datasets. These include:
\begin{itemize}
    \item \textbf{Weather:} Fog, rain, overcast skies, and varying visibility levels.
    \item \textbf{Lighting:} Low-light and uneven lighting conditions (e.g., shadowed roads).
    \item \textbf{Scenes:} Urban congestion, rural roads, flyovers, intersections, roadwork zones, and unmarked lanes.
\end{itemize}

This diversity is essential for developing object detection models that generalize across unpredictable and complex traffic environments, especially in developing regions.

\subsection{Data Annotation Guidelines}

Annotations in the \textit{DriveIndia} dataset were performed using strict, evidence-based protocols to ensure high-quality and consistent labeling. The key guidelines followed are:

\begin{itemize}
    \item \textbf{Bounding Box Policy:}
    \begin{itemize}
        \item Boxes were drawn tightly around the main object, excluding small extremities and accessories.
        \item Occluded objects were annotated only for the visible parts, with emphasis on key regions (e.g., heads/shoulders for humans, wheels for cycles).
        \item Riders were annotated separately from the vehicles they operate.
        \item Edge-cut or motion-blurred objects were labeled only if at least 50\% of the object was visible.
    \end{itemize}

    \item \textbf{Image Filtering:}
    \begin{itemize}
        \item Images with severe blur were filtered from validation and test sets.
        \item A Laplacian variance threshold of 20 was used to exclude low-clarity images \cite{blur_laplasian1}.
    \end{itemize}

    \item \textbf{Quality Assurance:}
    \begin{itemize}
        \item Two-stage verification: random sampling and senior review for annotation consistency.
        \item Automated checks ensured:
        \begin{itemize}
            \item Each image had a valid corresponding label file.
            \item Empty or invalid annotations were removed.
            \item Class distributions were monitored across train/test/val splits to avoid data leakage.
        \end{itemize}
    \end{itemize}

    \item \textbf{Design Motivation:}
    \begin{itemize}
        \item The guidelines were informed by recent research showing that tight and clean annotations directly improve object detection performance \cite{annotation2020}.
    \end{itemize}
\end{itemize}

\begin{figure*}[h]
    \centering
    \includegraphics[width=\linewidth]{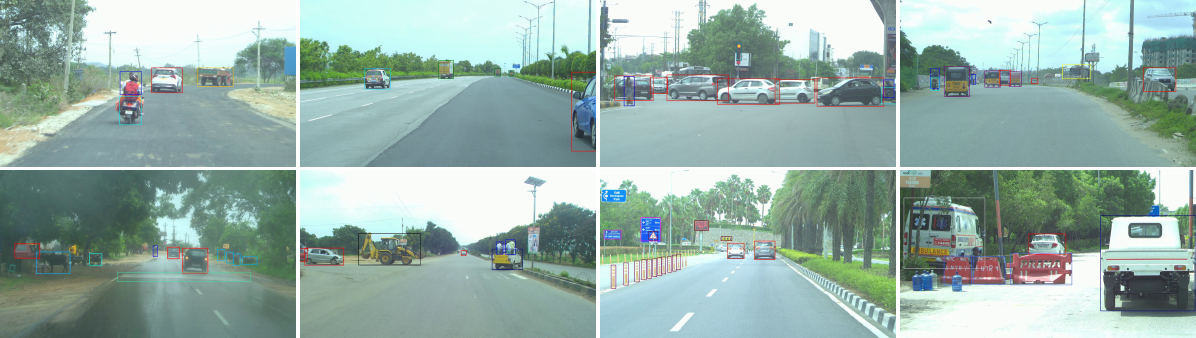}
    \caption{Example detections from the DriveIndia dataset, showcasing diverse object categories including vehicles, pedestrians, infrastructure elements, and environmental obstacles under varied conditions.}
    \label{fig:8 samples}
\end{figure*}

\subsection{Dataset Organization}
The \textit{DriveIndia} dataset is organized into standard training, validation, and test splits. The directory structure, shown below, separates images from their corresponding YOLO format annotation files.

\dirtree{%
.1 DriveIndia/.
.2 images/.
.3 train/.
.3 val/.
.3 test/.
.2 labels/.
.3 train/.
.3 val/.
.3 test/.
}

This organization ensures compatibility with standard object detection frameworks. Each image file has a corresponding \texttt{.txt} annotation file in the matching subdirectory.

\section{Benchmark: Object Detection}

To establish DriveIndia as a robust benchmark for object detection in uncertain real-world driving conditions, we evaluated several recent, high-performing detection models—including multiple YOLO versions \cite{yolo, yolov5, yolov8, yolov9, yolo11} and state-of-the-art transformer-based detectors such as RT-DETR \cite{RT-DETR} and EfficientDet \cite{efficientdetscalableefficientobject}. This section presents baseline results and discusses challenges uncovered by class-wise evaluation.

\subsection{Experimental Protocol}

All models were trained on the DriveIndia dataset with a standardized protocol: 54,856 images for training and 6,700 each for validation and testing. We adopted common hyperparameters across all runs for fairness. Model performance was assessed on the test set using widely adopted detection metrics: $mAP_{50}$, $mAP_{50{:}95}$, precision, and recall.

\subsection{Quantitative Results}

Table~\ref{tab:yolo_results} summarizes the results for all evaluated models. The YOLO family remains highly competitive, with YOLOv8 achieving the highest $mAP_{50}$ (78.7\%), while YOLOv9 offers the best precision/recall balance. Figure ~\ref{fig:8 samples} visualizes typical DriveIndia scenes by the YOLOv8 Model, reflecting the dataset’s real-world diversity.

\begin{table}[htbp]
\caption{Benchmark Results on DriveIndia}
\centering
\begin{tabular}{c|cc|cc}
\toprule
\textbf{Model} & $\mathbf{mAP_{50}}$ & $\mathbf{mAP_{50{:}95}}$ & \textbf{Precision} & \textbf{Recall} \\
\midrule
YOLOv5   & 0.767 & 0.545 & 0.778 & 0.725 \\
\rowcolor{gray!20} 
YOLOv8   & \textbf{0.787} & \textbf{0.564} & 0.775 & \textbf{0.754} \\
YOLOv9   & 0.779 & 0.560 & \textbf{0.791} & 0.734 \\
YOLOv11  & 0.717 & 0.506 & 0.714 & 0.660 \\
RT-DETR  & 0.765 & 0.553 & 0.764 & 0.728 \\
EfficientDet & 0.726 & 0.512 & 0.743 & 0.678 \\
\bottomrule
\end{tabular}
\label{tab:yolo_results}
\end{table}

\subsection{Per-Class Performance Analysis}

\begin{figure}
    \centering
    \includegraphics[width=\linewidth]{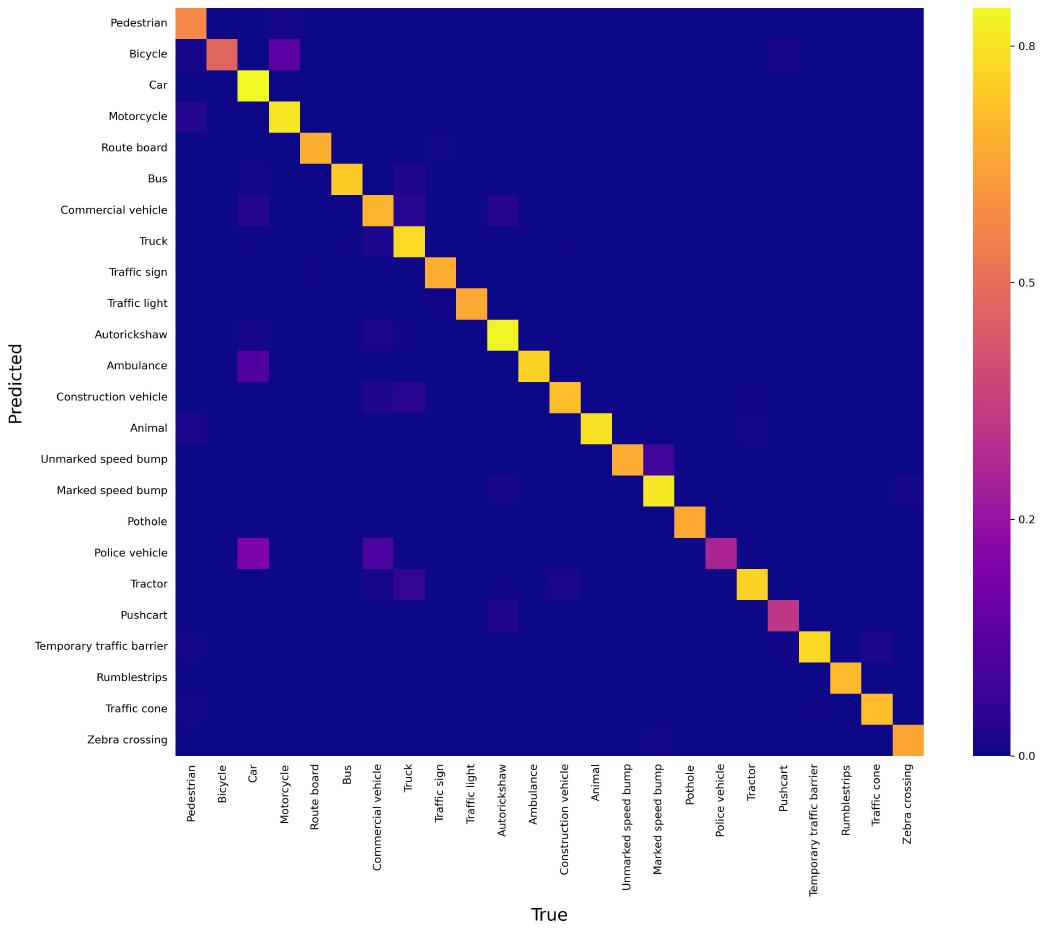}
    \caption{Confusion Matrix, highlighting confusion between visually or contextually similar categories.  }
    \label{fig:confuse}
\end{figure}

A detailed breakdown of class-wise metrics is presented in Table~\ref{tab:driveindia_classwise}. While common objects like \textit{car}, \textit{truck}, and \textit{bus} achieve high $mAP_{50}$ scores ($>0.7$), rare and small-scale classes such as \textit{pushcart}, \textit{pothole}, \textit{police vehicle}, and \textit{unmarked speed bump} show lower detection accuracy ($mAP_{50{:}95} < 0.5$). Figure~\ref{fig:confuse} displays the confusion matrix, highlighting frequent misclassifications, particularly among visually similar or underrepresented classes.

\begin{table}[ht]
\centering
\caption{Per-class detection performance on the DriveIndia `Validation set'. Color indicates \textcolor{green!70!black}{high}, \textcolor{orange!85!black}{moderate}, and \textcolor{red!75!black}{low} detection scores.}
\label{tab:driveindia_classwise}
\begin{adjustbox}{width=\columnwidth}
\begin{tabular}{@{}lrrcc@{}}
\toprule
\textbf{Class} & \textbf{Images} & \textbf{Instances} & $\mathbf{mAP_{50}}$ & $\mathbf{mAP_{50{:}95}}$ \\
\midrule
Pedestrian                    & 4,344 & 11,806 & \textcolor{green!70!black}{0.754} & \textcolor{green!70!black}{0.501} \\
Bicycle                       & 73 & 80  & \textcolor{orange!85!black}{0.645} & \textcolor{orange!85!black}{0.455} \\
Car                           & 4,911 & 11,959 & \textcolor{green!70!black}{0.945} & \textcolor{green!70!black}{0.819} \\
Motorcycle                    & 4,156 & 9,795 & \textcolor{green!70!black}{0.947} & \textcolor{green!70!black}{0.703} \\
Route board                   & 298 & 333  & \textcolor{green!70!black}{0.841} & \textcolor{green!70!black}{0.587} \\
Bus                           & 997 & 1,181 & \textcolor{green!70!black}{0.878} & \textcolor{green!70!black}{0.735} \\
Commercial vehicle            & 1,457 & 1,798 & \textcolor{green!70!black}{0.840} & \textcolor{green!70!black}{0.689} \\
Truck                         & 2,036 & 2,918 & \textcolor{green!70!black}{0.899} & \textcolor{green!70!black}{0.752} \\
Traffic sign                  & 684 & 873  & \textcolor{green!70!black}{0.851} & \textcolor{green!70!black}{0.565} \\
Traffic light                 & 297 & 432  & \textcolor{orange!85!black}{0.710} & \textcolor{orange!85!black}{0.353} \\
Auto-rickshaw                 & 2,291 & 3,242 & \textcolor{green!70!black}{0.940} & \textcolor{green!70!black}{0.803} \\
Ambulance                     & 20 & 20  & \textcolor{green!70!black}{0.780} & \textcolor{green!70!black}{0.659} \\
Construction vehicle          & 149 & 156  & \textcolor{orange!85!black}{0.735} & \textcolor{green!70!black}{0.515} \\
Animal                        & 62 & 135  & \textcolor{green!70!black}{0.769} & \textcolor{orange!85!black}{0.471} \\
Unmarked speed bump           & 37 & 41  & \textcolor{red!75!black}{0.454} & \textcolor{red!75!black}{0.182} \\
Marked speed bump             & 78 & 91  & \textcolor{orange!85!black}{0.715} & \textcolor{orange!85!black}{0.343} \\
Pothole                       & 189 & 358  & \textcolor{red!75!black}{0.497} & \textcolor{red!75!black}{0.232} \\
Police vehicle                & 10 & 10  & \textcolor{orange!85!black}{0.511} & \textcolor{orange!85!black}{0.470} \\
Tractor                       & 173 & 178  & \textcolor{green!70!black}{0.843} & \textcolor{green!70!black}{0.633} \\
Pushcart                      & 26 & 29  & \textcolor{red!75!black}{0.391} & \textcolor{red!75!black}{0.206} \\
Temporary traffic barrier     & 145 & 278  & \textcolor{green!70!black}{0.837} & \textcolor{green!70!black}{0.547} \\
Rumble strips                  & 355 & 431  & \textcolor{green!70!black}{0.825} & \textcolor{orange!85!black}{0.488} \\
Traffic cone                  & 181 & 567  & \textcolor{green!70!black}{0.800} & \textcolor{orange!85!black}{0.367} \\
Zebra crossing                & 159 & 196  & \textcolor{orange!85!black}{0.684} & \textcolor{orange!85!black}{0.339} \\
\bottomrule
\end{tabular}
\end{adjustbox}
\end{table}

\subsection{Discussion: Class-Specific Detection Challenges}

Our evaluation\footnote{Color categories for Table \ref{tab:driveindia_classwise} \textbf{For $mAP_{50:95}$}, \textcolor{green!60}{green} $>$0.50, \textcolor{orange!80}{orange} 0.30–0.50, \textcolor{red!80}{red} $<$0.30. For\textbf{ $mAP_{50}$}, \textcolor{green!60}{green} $\geq$0.75, \textcolor{orange!80}{orange} 0.50–0.75, \textcolor{red!80}{red} $<$0.50.} as seen on Table \ref{tab:driveindia_classwise} shows that common road users—such as \textit{car}, \textit{Auto-rickshaw}, \textit{truck}, and \textit{motorcycle}—are detected with high accuracy ($mAP_{50}$ above 0.9), reflecting model robustness for prevalent classes in Indian traffic.

However, detection is notably weaker for rare or visually ambiguous classes. \textit{Bicycle} exhibits lower performance  ($mAP_{50}$ 0.645), likely due to frequent occlusion and visual overlap with motorcycles. Similarly, classes like \textit{animal} are often confused or missed due to appearance variability and limited samples.

Critical but underrepresented categories—such as \textit{ambulance} ($mAP_{50}$ 0.78) and \textit{police vehicle} ($mAP_{50}$ 0.51)—also pose challenges, as these often lack distinct visual features and are seen infrequently. Infrastructure-related classes such as \textit{pothole} and \textit{unmarked speed bump} achieve the lowest mAP, underscoring the difficulty of detecting small or subtle road features.

These results emphasize the need for improved data augmentation and targeted model enhancements for ambiguous and rare, yet safety-critical, classes—key priorities for robust ADAS in real-world Indian traffic scenarios.

\section{Dataset Availability and Licensing}

The DriveIndia dataset will be publicly released for academic and non-commercial research via the official TiHAN-IIT Hyderabad dataset portal:
\[
\texttt{\url{https://tihan.iith.ac.in/TiAND.html}}
\]

The release will include:
\begin{itemize}
    \item 66,986 RGB images annotated in YOLO format
    \item 24 traffic-relevant object categories
    \item Pre-defined training, validation, and test splits
\end{itemize}
\section{Conclusion}

We presented DriveIndia, a comprehensive object detection benchmark designed to reflect the full complexity and diversity of Indian traffic environments. With nearly 67,000 annotated images spanning 24 object categories—including rare and safety-critical classes—DriveIndia uniquely captures the real-world challenges of heterogeneous road users, occlusions, and adverse weather conditions such as fog and rain.

DriveIndia is fully compatible with modern detection frameworks, including YOLOv5, YOLOv8, YOLOv9, YOLOv11, RT-DETR, and EfficientDET, facilitating rigorous comparisons across architectures. Our large-scale experiments reveal that even state-of-the-art detectors face significant drops in accuracy for rare, small, or ambiguous classes (e.g., pushcart, unmarked speed bump, bicycle), despite strong overall $mAP$ on common vehicles.

These results highlight persistent gaps in current object detection approaches and the importance of context-aware models, targeted data augmentation, and improved handling of long-tail categories. By releasing DriveIndia, we aim to accelerate research towards safer, more reliable ADAS and autonomous systems in unstructured, real-world scenarios.

We invite the research community to leverage and extend DriveIndia, and we welcome contributions that push the boundaries of robust visual perception in challenging traffic environments.

\section{Limitations and Future Research}

Despite the scale and diversity of DriveIndia, certain limitations remain that present opportunities for further advancement:

\begin{itemize}
    \item \textbf{Geographical Scope:} The dataset primarily covers southern Indian regions and does not capture the full diversity of terrains, traffic patterns, and signage found across other parts of India, such as hilly terrains or dense metro areas in the north and northeast.

    \item \textbf{Sparse Representation of Rare Classes:} Safety-critical but rare classes (e.g., \textit{ambulance}, \textit{police vehicle} have limited instances, challenging robust model training for long-tail categories.

    \item \textbf{Absence of Night-Time Scenes:} Although varied weather and lighting conditions are included, night-time scenarios are underrepresented, affecting generalization in low-light settings.

    \item \textbf{Single-Modality Limitation:} DriveIndia uses only RGB imagery. Incorporating additional modalities such as LiDAR would improve perception under adverse weather and complex road geometries.

    \item \textbf{Lack of 3D and Semantic Annotations:} The dataset includes only 2D bounding boxes. Adding semantic masks and 3D bounding boxes (or point-level LiDAR) would enable broader AV perception tasks.

    \item \textbf{Benchmark Scope:} Current evaluation is limited to object detection. Extending to tracking, segmentation, and domain adaptation would enhance research applicability for ADAS systems.
\end{itemize}

We envision DriveIndia as a foundational benchmark for robust, multimodal perception research on unstructured roads. Community contributions in these directions are encouraged.

\section*{Acknowledgements}

This research was supported by the DST National Mission on Interdisciplinary Cyber-Physical Systems (NM-ICPS) and TiHAN at IIT Hyderabad, whose funding and resources were instrumental.

I express my sincere gratitude to Aravind Kumapati and Menasath Santhosh for their diligent efforts in data collection and annotation, which ensured the development of high-quality datasets critical to this work. I also extend my thanks to the anonymous team members whose contributions to data processing and annotation were invaluable.

\bibliographystyle{IEEEtran}
\bibliography{mybibfile}

\end{document}